%% file: main.tex
\documentclass[11pt,letterpaper]{article}
\usepackage{amsmath,amssymb,amsthm}
\usepackage{deepthink}
\usepackage[round]{natbib}

\input{preamble_shared.tex}

\newcommand{\censuswraplines}{10}
\newcommand{\censuswrapwidth}{0.36\linewidth}

\title{Scaling and Distilling Text Embeddings for Better Diffusibility}

\newcommand{\aname}[2]{\if\relax\detokenize{#2}\relax\textbf{#1}\else\href{#2}{\textbf{#1}}\fi}
\authorblock{
  \aname{Zekai Zhang}{https://la0ka1.github.io/}\textsuperscript{1},
  \aname{Yunjie Tian}{https://sunsmarterjie.github.io/},
  \aname{Yanjin He}{https://yanjinhe.github.io/}\textsuperscript{1},
  \aname{Xiaoyan Zhang}{https://xiaoyanzhang1.github.io/}\textsuperscript{1},
  \aname{Dongdi Zhao}{},
  \aname{Qing Qu}{https://qingqu.engin.umich.edu/}\textsuperscript{1},
  \aname{Di Fu}{}
}
\affiliation{\textsuperscript{1}University of Michigan}
\abstracttext{\noindent\input{Sections/abstract}}
\date{September 2026}   %
\correspondence{zzekai@umich.edu, \{tianyunjie96, fu.burning\}@gmail.com}
\resources{\href{https://github.com/la0ka1/diffusing-scaled-text-embeddings}{Code} \textbar{} \href{https://la0ka1.github.io/diffusing-scaled-text-embeddings/}{Project page}}

\begin{document}
\makeDeepthinkHeader

\input{body}

\end{document}

%% file: preamble_shared.tex
\input{math_commands.tex}

\usepackage{url}
\usepackage{graphicx} 
\usepackage{booktabs}
\usepackage{amsmath}
\usepackage{amssymb}
\usepackage{multirow}
\usepackage{subcaption}
\usepackage{wrapfig}
\definecolor{elfnavy}{HTML}{0A3D77}
\usepackage{xcolor,graphicx,wrapfig,enumitem}

\definecolor{darkgreen}{RGB}{0,110,0}

\usepackage{tcolorbox}
\tcbuselibrary{skins}
\newtcolorbox{obsbox}{enhanced, colback=black!5!white, colframe=black!35!white, boxrule=0.4pt, arc=4pt, left=9pt, right=9pt, top=6pt, bottom=6pt,
  drop fuzzy shadow=black!45!white}
\definecolor{figpurple}{HTML}{9467BD}   %
\definecolor{sampnavy}{RGB}{31,64,104}
\definecolor{samporange}{RGB}{198,89,17}
\newtcolorbox{teacherbox}[1]{colback=sampnavy!4!white, colframe=sampnavy,
  title={\normalsize\bfseries #1}, fontupper=\normalsize\raggedright,
  boxrule=0.7pt, arc=2.5pt, left=7pt, right=7pt, top=5pt, bottom=5pt}
\newtcolorbox{studentbox}[1]{colback=samporange!5!white, colframe=samporange,
  title={\normalsize\bfseries #1}, fontupper=\normalsize\raggedright,
  boxrule=0.7pt, arc=2.5pt, left=7pt, right=7pt, top=5pt, bottom=5pt}

%% file: math_commands.tex
\usepackage{amsmath,amsfonts,bm}

\def\eqref#1{equation~\ref{#1}}
\def\Eqref#1{Equation~\ref{#1}}

\def\1{\bm{1}}

\DeclareMathAlphabet{\mathsfit}{\encodingdefault}{\sfdefault}{m}{sl}
\SetMathAlphabet{\mathsfit}{bold}{\encodingdefault}{\sfdefault}{bx}{n}

%% file: Sections/abstract.tex
Diffusion language models (DLMs) offer a promising alternative to autoregressive (AR) language generation. Recent advances in continuous DLMs, which apply latent diffusion to continuous text embeddings, raise a practical question: which embedding makes the best latent space, \textit{i.e.}, the most diffusible? To answer this, we search through different embeddings and find that \textbf{scaling} the embedding model to stronger ones within the same family (T5 to T5Gemma-1 to T5Gemma-2) greatly improves generative performance. But the raw T5Gemma-2 embeddings are still not optimal. They are so discriminative that even the embeddings of plausible alternative words are separated, which makes the generation vulnerable to imperfect sampling. Consequently, continuous diffusion often fails to reach any of them and ends up at an invalid embedding instead. To address this, we \textbf{distill} T5Gemma-2 into a student encoder that learns the teacher's decoded probabilities as soft labels. Learning from such soft labels makes the student pull the alternative embeddings closer while maintaining the encoding-decoding mechanism. The distilled embeddings form a more connected and diffusible latent space, improving over the vanilla T5Gemma-2 embeddings. As a result, our medium-sized DLM achieves \textbf{Gen.\ PPL 17.8} (against real-text PPL 15.4) at real-text entropy on OpenWebText, outperforming GPT-2-M on Gen.\ PPL.\

%% file: body.tex
\begin{figure}[!h]\centering
\includegraphics[width=\linewidth]{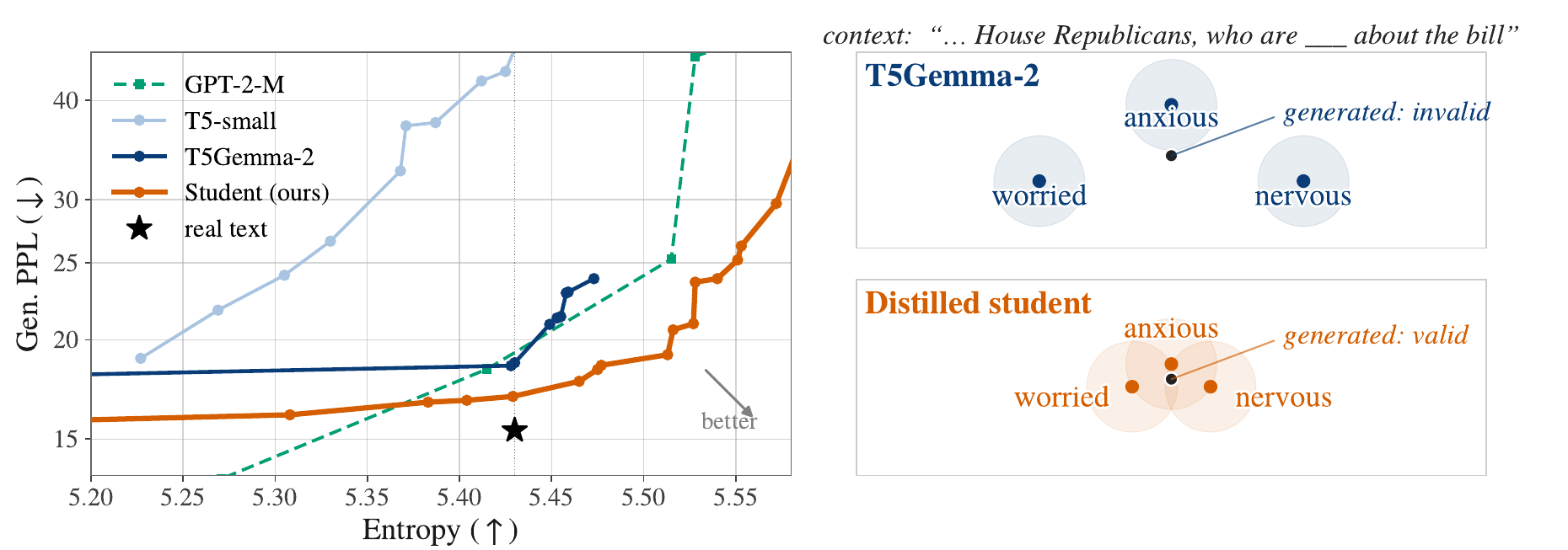}
\caption{\textbf{Overview of our performance and mechanism.} \emph{Left}: the same DLM trained on our embeddings (distilled from T5Gemma-2) reaches real-text entropy with the best quality. \emph{Right}: T5Gemma-2 maps plausible candidates apart in the embedding space, so a generated embedding often reaches none of those. Our distilled embedding pulls the candidates closer, forming a continuous region where an embedding at least decodes to one of them.
}
\label{fig:opening}
\end{figure}

\section{Introduction}

Diffusion models~\citep{ho2020denoising} have achieved huge success in continuous modalities such as image~\citep{flux-2-2025} and video~\citep{seed2026seedance}. A natural question is whether their strong modeling transfers to language generation, and whether Diffusion Language Models (DLMs) can serve as a more parallelizable and efficient paradigm beyond autoregressive (AR) generation. There are two families of DLMs: the first is discrete DLMs~\citep{austin2021structured, nie2025large}, which apply a discretized diffusion process directly on tokens, and use likelihood training~\citep{lou2023discrete}. They have drawn the most attention since they can be adapted from foundation models and scale with them~\citep{ye2025dream, team2026diffusiongemma}. The second is continuous DLMs~\citep{li2022diffusion} which we study in this paper. They apply latent diffusion on text embeddings and decode generated embeddings back to tokens, and are trained with MSE loss. Interestingly, though introduced as early as discrete DLMs, continuous DLMs remain less studied until their recent revival~\citep{hu2026elf, guo2026continuous}, partly due to the incorporation of new techniques including self-conditioning~\citep{chen2022analog}, diffusion transformers~\citep{peebles2023scalable, li2026back}, and the adoption of pretrained embeddings~\citep{meshchaninov2026cosmos}.

A central question for continuous DLMs is which text embedding to use as the latent space. The same question is already a major topic in image diffusion~\citep{yao2025reconstruction, bfl2025representation}, where the recent Representation Autoencoder (RAE)~\citep{zheng2026diffusion} finds that replacing the VAE latents~\citep{rombach2022high} with pretrained image embeddings improves diffusibility. For text, earlier work finds that pretrained embeddings improve continuous DLMs over simple embeddings from lookup tables~\citep{lovelace2023latent, zhang2023planner}. However, existing approaches either modify the embeddings and the diffusion model jointly~\citep{lemercier2026diladiff, li2026convergeflow} or rely on relatively older embedding models~\citep{hu2026elf}, making it difficult to isolate the contribution of the embedding space or fully realize its potential.

In this paper, we study the embedding space for continuous DLMs in isolation. We fix the diffusion side as the recent Embedded Language Flow (ELF)~\citep{hu2026elf} framework for its minimalist design, and change only the embedding. We find two things. \textbf{(i) Scaling the embeddings leads to a more diffusible latent space}. By scaling, we mean moving to stronger embedding models of the same family, which are pretrained with more data and better recipes, and not only adding parameters. We find that along with the scaling of T5-small/base~\citep{raffel2020exploring} $\to$ T5Gemma-1 $\to$ T5Gemma-2~\citep{zhang2025t5gemma}, diffusibility also increases, and replacing the default T5-small embedding in ELF with T5Gemma-2-270M reduces generative perplexity by about 40\% at the same entropy. \textbf{(ii) Scaled embeddings can be further enhanced via distillation.} A student encoder trained to match the probabilities from the T5Gemma-2 decoder is more diffusible than the teacher. As a result, the same DLMs trained on the distilled embeddings improve over the teacher on perplexity and MAUVE and sample more stably, as illustrated in Fig.~\ref{fig:opening}.

Distillation helps because it fixes a problem of the raw T5Gemma-2 embeddings, which is that they are hard to generate perfectly. When sampling is insufficient or imperfect, the trajectories end at invalid embeddings that deviate from the real ones and decode to an uncertain or even wrong word~\citep{shabalin2026gaussian}. We relate this to the strong representation learning of T5Gemma-2, which makes its embeddings information-dense and distinct, so even the words that could fill the same position are kept apart~\citep{yu2020learning}. That helps discriminative tasks but hurts generation, since a continuous trajectory has to settle on one of several separate candidates and can end at none of them. The soft labels in distillation mitigate such separation, since the student learns to predict the plausible candidates so the alternative embeddings are placed near each other, and the generated embedding lies in the region they form.

Our results also shed light on new understandings of language representation learning and generation. Although text embeddings are considered contextualized and continuous, they inherit discreteness from text and do not fill the entire embedding space, so continuous DLMs can still generate embeddings that do not represent any word. This, in turn, is the other side of continuous language generation. Though more flexible than predicting tokens, they also risk deviating and generating invalid embeddings. Whether continuous DLMs will become a fully workable route still needs to be tested on even larger and practical scales, and the scaling and distillation here provide a basis for that. In all, our contribution can be summarized as follows:
\begin{itemize}[leftmargin=*]
   \item We show that scaled embeddings make good latent spaces for continuous DLMs, providing a strong baseline and starting point for continuous DLMs with T5Gemma-2 embeddings.
   \item We study the diffusibility of text embeddings from the perspective of generating valid, decodable embeddings.
   \item We enhance T5Gemma-2's diffusibility by distilling its decoder probabilities, providing insights for taming and incorporating foundation models for continuous DLMs.
\end{itemize}

\section{Preliminary and Related Work}
\input{Sections/prelim}

\section{Scaling Text Embedding for Continuous DLMs}
\input{Sections/scaling}

\section{Enhancing the Diffusibility via Distillation}\label{sec:distill}
\input{Sections/distilling}

\section{Main Experiments}\label{sec:exp}
\input{Sections/experiments}

\section{Conclusion}
\input{Sections/conclusion}

\section*{Ethics statement}
We use fully public corpora (OpenWebText, LM1B, WikiText-103) and models (T5Gemma-2, T5-small, ModernBERT). No human subjects or private data are involved. Our models generate unfiltered web-style text that can be false or biased; we use the samples only for evaluation.

\section*{Reproducibility statement}
Our continuous DLM code is adapted from the public ELF-pytorch implementation. We change only the embedding model and keep the training and sampling recipe (App.~\ref{app:train}). Distillation details are in App.~\ref{app:distill}; App.~\ref{app:eval} gives the evaluation protocol, the sampling grid, and how each table row is chosen. Baselines are re-run with their authors' code and checkpoints, or cited where no checkpoint is public. Generated samples are in App.~\ref{app:examples}.

\section*{Disclosure of AI use}
We use generative AI tools to implement methods (training, sampling, and evaluation codes), to help design and give feedback on experiments, to clean and reformat datasets, to search for additional literature, and to support qualitative analysis of generated samples. We have not used them to generate synthetic datasets, to formulate mathematical claims, to write proofs, or to assist with translation; the other required disclosures are not applicable. Additionally, we use generative AI tools to create and edit figures, to draft parts of the text (as suggestions; the final text is written by the authors). The research ideas, including scaling the embeddings and distilling them, originate from the human authors; AI tools are given direct, clear instructions and refined over multiple turns. All AI-assisted code is run and checked by the authors, and the results are re-checked against the external baselines. We take responsibility for the final content of this work, including text, claims, or artifacts produced with the aid of generative AI.

\newpage
\bibliographystyle{iclr2027_conference}
\bibliography{ref.bib}

\appendix

\input{Sections/appendix}

%% file: Sections/prelim.tex
\subsection{Preliminaries}
Continuous DLMs are latent diffusion on text embeddings. We follow the ELF~\citep{hu2026elf} setup and first map the input token sequences $\bm{s}\in \mathcal{V}^L$ to the sequences of embeddings with a (frozen) embedding model:
\begin{align*}
    \bm{x} = \text{Emb}(\bm{s}) \in \mathbb{R}^{L\times d}
\end{align*}
This is also the only part we change throughout the paper.\footnote{Strictly speaking, we also change the final LM head since different embeddings have different vocab sizes.} Training and sampling are standard diffusion, where we train a denoiser $\bm{x}_\theta(\bm{z}_t, t)$ on embedding sequences corrupted by sequence-wise Gaussian noise $\bm\epsilon \in \mathbb{R}^{L\times d} $,
\begin{align*}
    \mathcal{L} = \mathbb{E}_{t, \bm{x}, \bm{\epsilon}} \Big[\frac{1}{(1-t)^2}\|\bm{x}_\theta(\bm{z}_t, t) - \bm{x}\|^2\Big], \quad \bm{z}_t := t\bm{x}+(1-t)\bm{\epsilon},
\end{align*}
and the optimal denoiser is the posterior mean:
\begin{align}
\bm x^\star(\bm z_t, t) = \mathbb{E}[\bm x \mid \bm z_t] = \sum_i p(\bm x_i \mid \bm z_t)\, \bm x_i , \label{eq:posterior}
\end{align}
where $\{\bm x_i\}$ are the embeddings of the training data, and weights $p(\bm x_i\mid\bm z_t)$ decrease with $\|\bm z_t-t\,\bm x_i\|$.

We also train a decoder that maps the generated embeddings back to tokens. It shares weights with the denoiser $\bm{x}_\theta$ plus a trainable head, and is trained on embedding sequences with token-wise corruption, where $\bm\lambda\in[0,1]^L$ is a per-token noise level
\begin{align}
    \mathcal{L}_{\text{CE}} = \text{CE}(\text{Head}_{\theta'}(\bm{x}_\theta(\tilde{\bm{z}}, t=1)), \bm{s} ), \quad  \tilde{\bm{z}} = \bm{\lambda}\,\bm{x} + (\bm{1}-\bm{\lambda})\,\bm{\epsilon}
\label{eq:loss_decoder}
\end{align}
This improves robustness for decoding imperfect embeddings. We also keep specific design choices of ELF (noise scheduling, self-conditioning, SDE sampling, etc.) unchanged; see App.~\ref{app:train}.

\subsection{Related work}
\paragraph{Continuous DLMs and their latents.} Earliest continuous DLMs adopt simple and co-trained embeddings~\citep{li2022diffusion}. However, co-training collapses the embeddings, since one point for all is the easiest way to reduce the MSE loss~\citep{dieleman2022continuous, gao2024empowering}. Recent work finds that pretrained text embeddings perform better~\citep{meshchaninov2026cosmos,hu2026elf}; this paper follows this line and tests their full potential.

\paragraph{Discrete DLMs.} Discrete DLMs use a generalized diffusion process with categorical corruption on the tokens~\citep{lou2023discrete,ou2024your, sahoo2024simple}. Importantly, they can be adapted from AR foundation models and benefit from their progress~\citep{gong2025scaling, team2026diffusiongemma, zhu2026flare}. We show that continuous DLMs can co-evolve with them too~\citep[see also][]{yang2026continuous}, by using embedding models that are adapted from such foundation models. There are also papers showing discrete DLMs can be enhanced via continuous text embeddings~\citep{zhou2025coevolutionary, lemercier2026diladiff}, or continuous relaxations~\citep{deschenaux2026simplexdiffusionmodels}.

\paragraph{Diffusibility of embeddings.} Which latent is easy to diffuse is a central question in image diffusion~\citep{chen2025dc, skorokhodov2025improving, xu2026making}, and replacing VAE latents with pretrained representations such as DINOv2~\citep{oquab2023dinov2} improves diffusibility~\citep{zheng2026diffusion, shi2026latent}. We find the text counterpart, where the scaled T5Gemma-2 provides a strong latent space, and instead of explicitly redesigning~\citep{zhang2023planner, jiang2026textldm} the latent, we distill the scaled encoder for diffusibility.

%% file: Sections/scaling.tex
\begin{obsbox}
\textbf{(1) Scaled embeddings are high-potential latents.} Embeddings trained with richer data and a better recipe can expose more information for the diffusion to learn, and scaling from T5-small to T5Gemma-2 cuts Gen.\ PPL by about $40\%$ at matched entropy.

\textbf{(2) Scaled embeddings can still be hard to generate.} Generated embeddings can be invalid and far from real ones. This is because T5Gemma-2 keeps the plausible words for a position apart, so a generated embedding can miss all of them and decodes problematically.
\end{obsbox}

\subsection{Scaling improves the diffusibility of embeddings}
\label{subsec:existing_embd}

We start with off-the-shelf embedding models, on which we train the same ELF-B on a 25\% subset of OpenWebText~\citep{Gokaslan2019OpenWeb} (512 tokens per sequence, denoted as `OWT-512') for faster comparison. We mainly study the family of encoder-decoder~\citep{sutskever2014sequence} models, including T5-small~\citep{raffel2020exploring} used in ELF, T5-base, T5Gemma-1-S/B~\citep{zhang2025encoder}, T5Gemma-2-270M~\citep{zhang2025t5gemma} (denoted as T5Gemma-2 throughout the paper); and we also include a strong modern encoder-only model, ModernBERT~\citep{warner2025smarter}. The embeddings are normalized by their global mean and variance, as in ELF.

\begin{figure}[t]\centering
\includegraphics[width=0.8\linewidth]{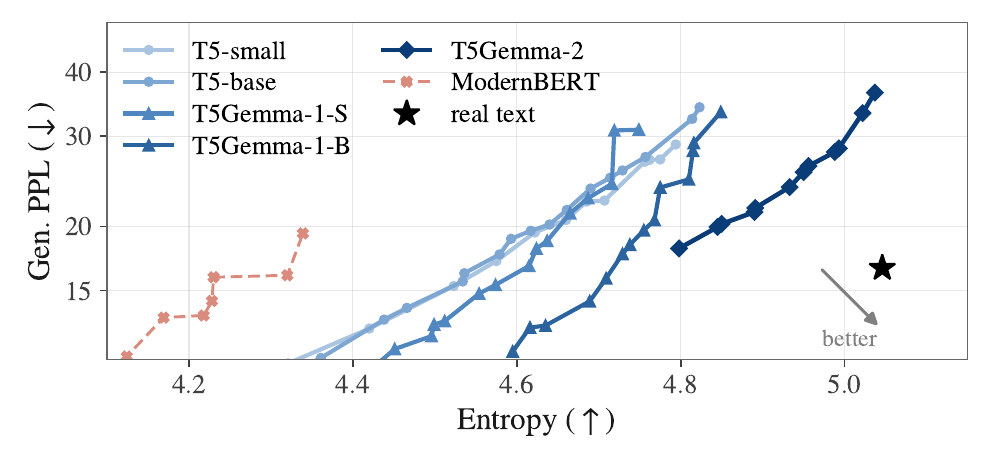}
\caption{ \textbf{Performance of the same ELF-B trained with different encoders on OWT-512.} T5Gemma-2 reaches the entropy of real text, and its Gen.\ PPL is about $40\%$ lower than T5-small's at the same entropy. Within the T5 family, T5Gemma-2 $>$ T5Gemma-1 $>$ T5-small.
}\label{fig:panel}
\end{figure}

We find that as the embedding models scale\footnote{By `scaling' we mean the overall pretraining scale of the embedding model (data, compute and recipe), not the encoder's parameter count alone.} from T5-small through T5Gemma-1 to T5Gemma-2, their diffusion performance (shown in the PPL-entropy curves in Fig.~\ref{fig:panel}) also improves. T5Gemma-1-B can outperform T5-small, while T5Gemma-2 further improves over T5Gemma-1-B. This suggests that stronger representation learning exposes richer information from the training data for the diffusion to learn.
While the encoder-only ModernBERT also performs well on downstream tasks, it is less suitable for diffusion and generates repetitive sentences. So an embedding that is strong on discriminative tasks is not necessarily a diffusible one.

Here, T5Gemma-2 is the only embedding space that reaches real-text entropy, and it cuts Gen.\ PPL by about 40\% at the same entropy as T5-small. We therefore pick T5Gemma-2 as the baseline for the rest of the paper. Here we focus on smaller-sized embedding models for a controllable experiment size, so further scaling to bulky encoders (e.g., T5Gemma-2-1B/4B) is left for future work.

\subsection{Scaled embeddings are hard to generate perfectly}\label{subsec:invalid}

However, when we inspect the generated embeddings, we find a failure mode: a generated embedding can be far from every candidate word. We take the candidates of a position from the deployed decoding head in \eqref{eq:loss_decoder}, since its top-$k$ are the words it would decode to, usually plausible alternatives for this position.\footnote{The pretrained T5Gemma-2 decoder picks similar words; see App.~\ref{app:aragree}.} To get the embedding of a candidate, we put it at this position of the decoded sequence and encode the sequence again. Embeddings that deviate from their candidates can still be decoded, but often with low confidence or to a wrong word.

\edef\savedintextsep{\the\intextsep}\setlength{\columnsep}{14pt}\setlength{\intextsep}{2pt}
\begin{wrapfigure}[\censuswraplines]{r}{\censuswrapwidth}\vspace{-4pt}\centering
\includegraphics[width=\linewidth]{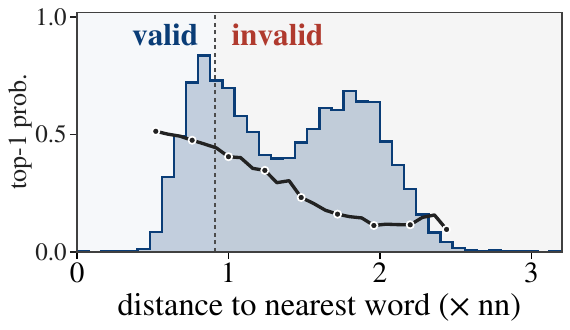}
\vspace{-14pt}
\caption{\textbf{Generated embeddings deviate from real T5Gemma-2's.}
}
\label{fig:census-teacher}
\end{wrapfigure}
\setlength{\columnsep}{10pt}

\begin{figure}[!htb]\centering
\includegraphics[width=\linewidth]{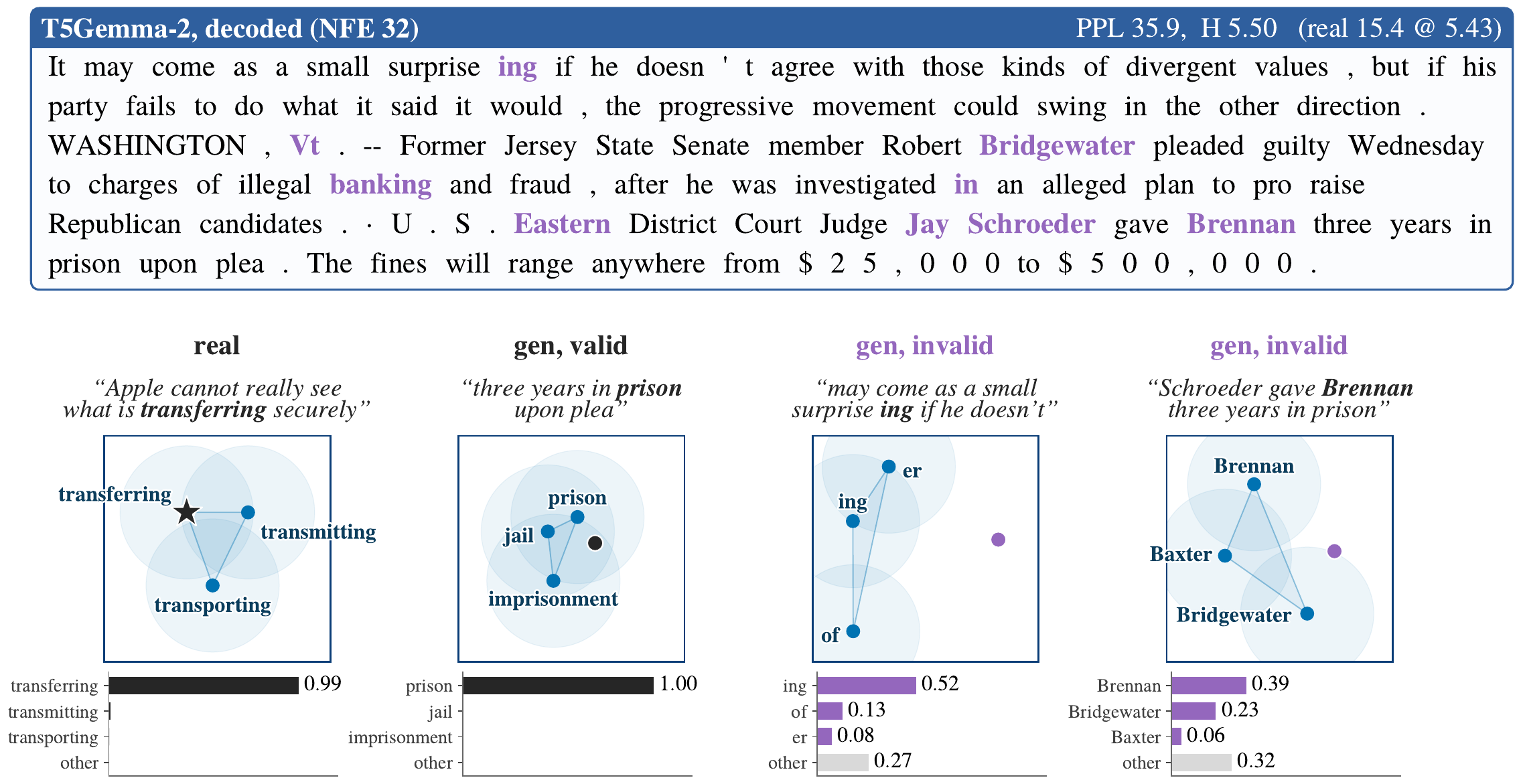}
\caption{\textbf{Visualizing the embedding error in 2D.} \emph{Top:} a sample generated at NFE $32$; uncertain tokens (top-1 below $0.6$) in \textcolor{figpurple}{purple} coincide with the grammar mistakes. \emph{Bottom:} four positions in the plane of their candidate words and the decoded probabilities. The disk marks the $0.91$\,nn range of a candidate, and an invalid one lands outside every disk and is decoded problematically. Distances are computed in the original $640$-dimensional space; the 2D layout is a distance-preserving map for display. See details in App.~\ref{app:maps}.
}
\label{fig:crystal}
\end{figure}

To quantify this, we count how often such embedding errors occur over $64$ generated sequences (Fig.~\ref{fig:census-teacher}).
We call a generated embedding \emph{invalid} when no candidate lies within reach. Formally, with $\bm x_1,\dots,\bm x_k$ the embeddings of the top-$k$ candidates, $\hat{\bm x}$ is invalid iff $\min_i\|\hat{\bm x}-\bm x_i\|>0.91\,\mathrm{nn}$, where nn is the median nearest-neighbor distance between real embeddings; we also flag positions with top-1 below $0.6$ as \emph{uncertain}. Both thresholds ($0.91$-nn, $0.6$) are derived from real text references, where the top-1 decoded probabilities from real embeddings are $\ge 0.98$, and the nearest neighbor distance is $1$\,nn by definition. So an embedding within $0.91$\,nn of a candidate is closer than the nn distance between real embeddings, and a top-1 below $0.6$ means the decoder does not produce a clear majority word. The exact threshold is not critical, as long as it divides the teacher's bimodal distances. On average, a generated sequence of $1024$ tokens has $142$ uncertain positions, and $109$ of them are invalid. For each uncertain position, we collect the distance from the generated embedding to its nearest candidate and the decoder's top-1 probability, and Fig.~\ref{fig:crystal} further shows one generated sequence in 2D.  We analyze at NFE $32$ for Sec.~\ref{subsec:invalid} and Sec.~\ref{subsec:easier}, as it is where the failure is most obvious. Our distillation is designed to fix this failure, and it also improves the generation at every NFE we test, from $64$ to $512$ in Sec.~\ref{sec:exp}.
\setlength{\intextsep}{\savedintextsep}   %

Previous works have studied such embedding errors as interpolation effects~\citep{chen2026interpolation, ashiq2026ddim, he2026diffusion} or mode-averaging~\citep{aithal2024understanding}, and a common finding is that discrete and separated targets are hard for continuous diffusion to generate~\citep{shabalin2026gaussian, zhang2026generalization}. We reiterate here that (ideally) sampling is guided by the optimal denoiser~\Eqref{eq:posterior}, which is a weighted mean of candidate embeddings. If candidates are separated, the mean falls between them and provides an ambiguous direction, and an imperfect sampling trajectory can end at an invalid location far from any of them. The decoder is then forced to predict with such embeddings, causing errors in the final generation.

%% file: Sections/distilling.tex
\begin{obsbox}
\textbf{(3) Distilling with soft labels makes the embeddings easier to generate.} A student that learns the teacher's decoded probabilities can place the plausible words closer to each other; the diffusion then lands on concrete words more often and generates better.
\end{obsbox}

\subsection{Distilling T5Gemma-2's Decoded Probabilities}
Why do T5Gemma-2 embeddings have the aforementioned separation problem? We relate it to how T5Gemma-2 is trained. As an encoder-decoder model, it mostly predicts and reconstructs its input as one-hot labels. This maximizes the likelihood of the input token against every alternative, pushing its embeddings away from semantically similar candidates that could fill its place~\citep{wu2024linguistic}. This helps with learning rich and distinctive representations~\citep{yu2020learning}, but hinders similar embeddings from forming a region and leads to invalid embeddings, as a trade-off between discrimination and generation.

\begin{figure}[h]\centering
\includegraphics[width=0.9\linewidth]{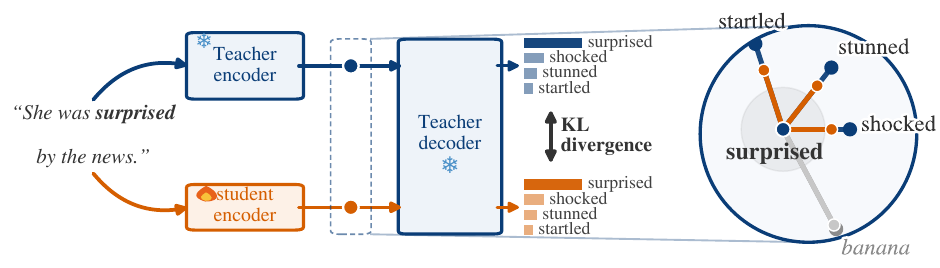}
\caption{\textbf{Illustration of the distillation pipeline and the effect on the embeddings.} \emph{Left:} The student encoder is trained to match the teacher's decoded probabilities (soft labels). \emph{Right:} The student pulls the embeddings of plausible candidates closer while keeping the irrelevant word far. The right panel is an illustration; measurements on real embeddings are in App.~\ref{app:realemb}.
}
\label{fig:distill_props}
\end{figure}

Therefore, we distill the T5Gemma-2 encoder into a student encoder that learns the teacher decoder's probabilities as soft labels (Fig.~\ref{fig:distill_props}). The soft labels spread over the plausible candidates of a position, so the student places these candidates closer~\citep{hinton2015distilling, muller2019does}. For a training sequence $\bm s\in\mathcal V^L$, we minimize
\begin{align*}
   \mathcal L_{\mathrm{KD}}(\theta)=\sum_{j=1}^L \mathrm{KL}\big(p^{\mathrm T}_j\,\|\,p^{\mathrm S}_j\big), \quad p^{\mathrm T} = \text{Dec}^{\mathrm T}(\text{Enc}^{\mathrm T}(\bm s)),\quad p^{\mathrm S} = \text{Dec}^{\mathrm T}(\text{Enc}^{\mathrm S}_\theta(\bm s))
\end{align*}
where $p^{\mathrm T}_j$ and $p^{\mathrm S}_j$ are the frozen teacher decoder's distributions at position $j$. We distill on the full OWT, initialize the student with $9$ evenly spaced layers from the teacher's $18$ layers~\citep{sanh2019distilbert}, and keep the top-$256$ candidates of the $262$k vocabulary with renormalized probabilities; see App.~\ref{app:distill}.

\paragraph{Ablations.} For the objective, replacing the KL on soft labels with CE on the one-hot input or MSE on the teacher's embeddings leads to students that never reach real-text entropy $5.43$ (CE$\leq 4.18$, MSE$\leq 5.26$, see Fig.~\ref{fig:mse_front_n1024} in Appendix). For the layer choice, we find that students initialized from the full $18$ layers to the $9$ we use consistently match or outperform the teacher, so the improvement is not due to reduced capacity. A randomly initialized student cannot converge. See App.~\ref{app:ablation} for details.

\paragraph{Generalizability.} Though we only distill on OWT, the student does not collapse to it. It is still able to encode and decode other text sources, e.g., WikiText~\citep{merity2016pointer}, as shown in App.~\ref{app:downstream}. As expected, the distillation costs some discriminative power and degrades downstream classification performance ($89.4$ to $78.1$ on SST-2~\citep{wang2018glue}), but it increases the diffusibility as we show in the following section.

\subsection{Distilled embeddings are easier to generate}\label{subsec:easier}

\begin{figure}
\centering
\includegraphics[width=\linewidth]{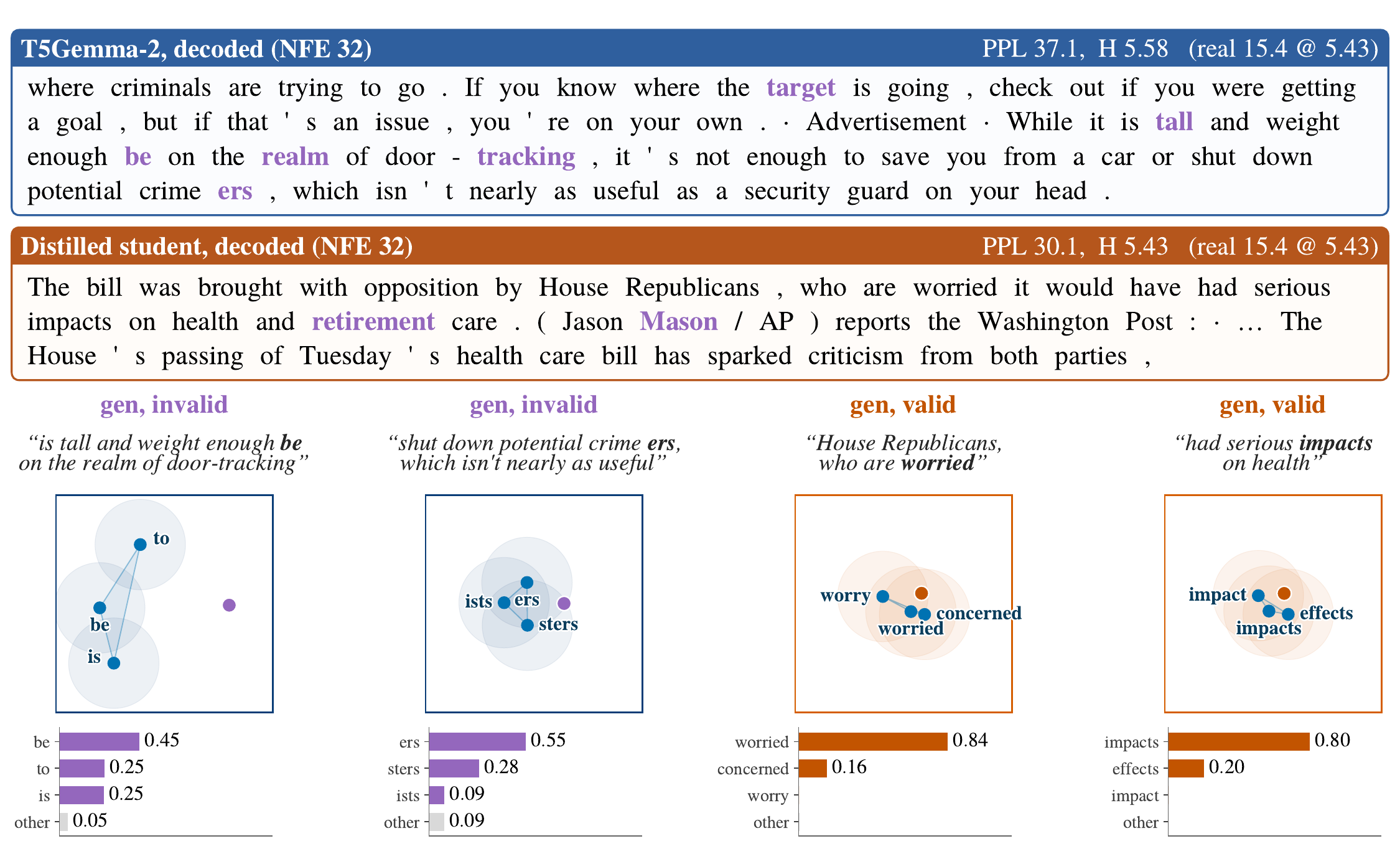}
\caption{\textbf{Distilled embeddings are more diffusible with lower embedding error.} Same layout as Fig.~\ref{fig:crystal}. The teacher has more uncertain embeddings (in \textcolor{figpurple}{purple}) which are mostly invalid, while the student's generated embeddings lie in the region formed by candidates that are close to each other. Over $64$ sequences, the teacher has $142$ uncertain positions per sequence, $109$ of which are also invalid; the student $42$ and $12$; see App.~\ref{app:census}.
}\label{fig:student_mitigation}
\end{figure}

To verify that the distilled student is more diffusible, we compare the embeddings generated by ELF-B trained on the teacher and student embeddings in Fig.~\ref{fig:student_mitigation}. By inspecting the generated samples at NFE $32$, we find that the teacher sample contains more invalid and uncertain embeddings than the student, along with more grammatical mistakes.

\begin{figure}[htb]\centering
\includegraphics[width=\linewidth]{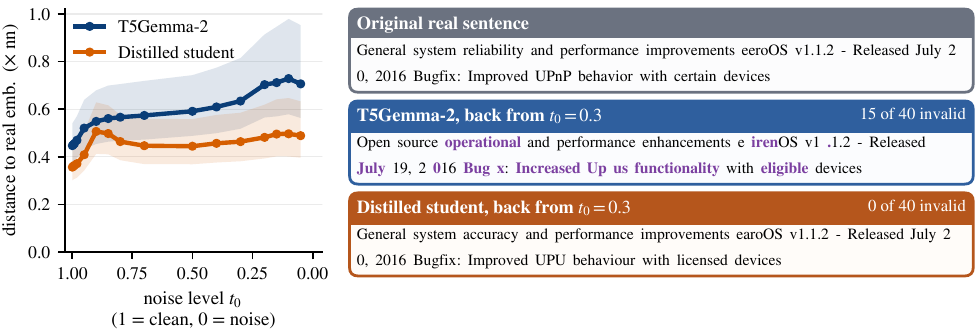}
\caption{\textbf{Round-trip denoising from the same sentence.} \emph{Left:} distance from each denoised embedding to the real embedding of the word it decodes to. \emph{Right:} one sequence denoised from $t_0=0.3$; invalid tokens in \textcolor{figpurple}{purple}.
}
\label{fig:roundtrip}
\end{figure}

We also test the denoising behavior of the diffusion trained on T5Gemma-2 vs. the student embeddings via round-trip denoising~\citep{meng2021sdedit}, as in Fig.~\ref{fig:roundtrip}. Given the same sentence and the same added noise, we see that the teacher has a much larger embedding error after denoising and changes the sentence, while the student returns valid embeddings closer to the original sentence, with meaningful fluctuations.

%% file: Sections/experiments.tex
We evaluate the diffusibility of our embeddings by training the same ELF-B and ELF-M models on OWT-1024 (the full OWT partitioned into 1024-token sequences) and on LM1B~\citep{chelba2013one} (128-token sequences). For each embedding+diffusion, we sweep the self-conditioning scale and NFE, re-tokenize the samples with the GPT-2 tokenizer\footnote{One problem is that, as different embeddings come with different tokenizers, the 1024-token sequences may not be directly comparable. So we re-tokenize everything with GPT-2 here. In App.~\ref{app:tokeq}, we further show that the tokenization effect is negligible and they are comparable.} and report the best Gen.\ PPL under GPT-2-Large at or above real-text entropy, and the mean and std are calculated over $3\times1024$ samples from 3 seeds.

\subsection{Generation Performance on OWT-1024}
We compare our performance against previous state-of-the-art continuous DLMs. We include GPT-2-S/M as AR baselines, and also Duo~\citep{sahoo2025diffusion} as a uniform-state discrete DLM representative. We find that ELF models trained on the T5Gemma-2 embeddings are already competitive, and our distilled embeddings further improve over the teacher, outperforming previous methods on Gen.\ PPL at real-text entropy. Note that the pretrained encoders have seen far more text than the OWT-trained baselines and the original T5-small, so rows with different encoders are not fully controlled; the controlled comparison in Table~\ref{tab:owt} is T5Gemma-2 against the student.

\begin{table}[t]
\centering
\caption{\textbf{Generation quality on OWT-$\mathbf{1024}$.} Best Gen.PPL at or above real-text entropy over the sampling sweep in App.~\ref{app:eval}; \textbf{bold} = best, \underline{underline} = second best. $^{\dagger}$: cited. $^{a}$: generation-time parameters, trunk$_{+\text{head}}$. $^{b}$: MAUVE is protocol-sensitive and serves as a secondary metric. \textcolor{gray}{Gray} rows fail to reach real-text entropy.
}
\label{tab:owt}
\setlength{\tabcolsep}{3pt}\small
\begin{tabular}{l|cccc}
  \toprule
  Method & Gen.\#Params$^{a}$ & Gen.PPL($\downarrow$) & Entropy($\uparrow$) & MAUVE$^{b}$($\uparrow$) \\
  \midrule
  \midrule
  Real Text & --- & 15.4$\pm$0.3 & 5.43$\pm$0.01 & 0.95$\pm$0.00 \\
  \midrule
  \midrule
  \multicolumn{5}{l}{\emph{Base scale}} \\
  GPT-2-S (AR)~\citep{radford2019language} & $85_{+39}$M & \underline{34.1$\pm$0.8} & 5.45$\pm$0.02 & \underline{0.88$\pm$0.02} \\
  Duo~\citep{sahoo2025diffusion} & $92_{+77}$M & 45.6$\pm$0.5 & 5.44$\pm$0.00 & \textbf{0.95$\pm$0.02} \\
  AURORA-LM$^\dagger$~\citep{liang2026aurora} & 130M & 98.4$\pm$1.4 & 5.45$\pm$0.01 & 0.85$\pm$0.03  \\
  LDLM$^{\dagger}$~\citep{meshchaninov2026train} & $132_{+66}$M & 34.3$\pm$0.3 & 5.46$\pm$0.01 & 0.31$\pm$0.05 \\
  \textcolor{gray}{CoBit-S}~\citep{batzolis2026cobit} & \textcolor{gray}{$128_{+6}$M} & \textcolor{gray}{37.3$\pm$0.5} & \textcolor{gray}{5.33$\pm$0.00} & \textcolor{gray}{0.69$\pm$0.03} \\
  \textcolor{gray}{T5-small + ELF-B}~\citep{hu2026elf} & \textcolor{gray}{$89_{+16}$M} & \textcolor{gray}{52.1$\pm$3.2} & \textcolor{gray}{5.31$\pm$0.02} & \textcolor{gray}{0.41$\pm$0.04} \\
  \textcolor{gray}{T5Gemma-2 + ELF-B} & \textcolor{gray}{$89_{+168}$M} & \textcolor{gray}{38.5$\pm$0.5} & \textcolor{gray}{5.41$\pm$0.01} & \textcolor{gray}{0.76$\pm$0.05} \\
  \textbf{Student (ours) + ELF-B} & $89_{+168}$M & \textbf{31.2$\pm$0.2} & 5.44$\pm$0.00 & 0.86$\pm$0.01 \\
  \midrule
  \midrule
  \multicolumn{5}{l}{\emph{Medium scale}} \\
  GPT-2-M (AR) & $302_{+52}$M & 20.8$\pm$0.3 & 5.45$\pm$0.00 & \textbf{0.91$\pm$0.01} \\
  CoBit-M & $453_{+9}$M & 31.3$\pm$1.1 & 5.43$\pm$0.01 & 0.49$\pm$0.03 \\
  \textcolor{gray}{T5-small + ELF-M} & \textcolor{gray}{$327_{+16}$M} & \textcolor{gray}{32.8$\pm$0.3} & \textcolor{gray}{5.36$\pm$0.01} & \textcolor{gray}{0.86$\pm$0.02} \\
  T5Gemma-2 + ELF-M & $327_{+168}$M & \underline{19.3$\pm$0.2} & 5.44$\pm$0.00 & 0.86$\pm$0.02 \\
  \textbf{Student (ours) + ELF-M} & $327_{+168}$M & \textbf{17.8$\pm$0.1} & 5.45$\pm$0.01 & \underline{0.89$\pm$0.01}  \\
  \bottomrule
  \end{tabular}
\end{table}

\begin{figure}[t]\centering
\includegraphics[width=\linewidth]{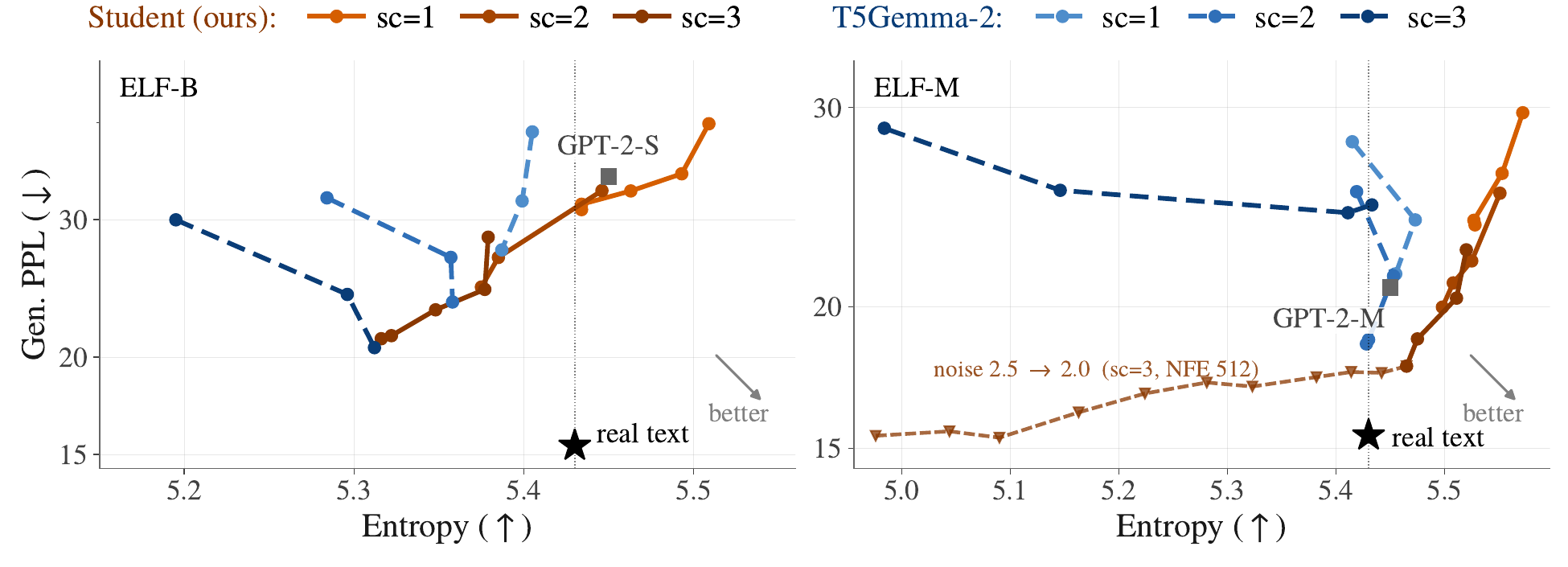}
\caption{\textbf{Sampling ablation of ELF-B (left) and ELF-M (right) models.} Each curve stands for different NFEs (64--512) under the same self-conditioning. The dotted curve on the right varies the sampler's noise scale. Our embedding consistently outperforms raw T5Gemma-2.}
\label{fig:sampling_abl}
\end{figure}

\paragraph{Sampling ablations.} We visualize the sampling sweeps in Fig.~\ref{fig:sampling_abl} that lead to Table~\ref{tab:owt}. Across different sampling configurations, the student consistently reaches lower Gen.\ PPL at higher entropy than the T5Gemma-2 teacher. Notably, T5Gemma-2 collapses at extremely low NFEs such as 16--32, due to under-integration (not shown in Fig.~\ref{fig:sampling_abl}), while our distilled embeddings are more stable.\footnote{In fact, at NFE 64 T5Gemma-2 can still collapse to a language mix, see App.~\ref{app:examples} for the collapsed samples.}

\subsection{Generation Performance on LM1B}
We report the performance on LM1B ($128$-token sequences) in Table~\ref{tab:lm1b} to show that the improvement is not due to overfitting or familiarity with OWT data, but a better geometry. The distilled student again performs well and surpasses the raw T5Gemma-2 embeddings on PPL. LM1B is a problematic corpus~\citep{pynadath2026generative} and serves as a check rather than a comparison, as we can see the re-trained AR model even has a lower PPL than real text.

\begin{table}[!htb]
\centering
\caption{\textbf{Generation quality on LM1B.} Same protocol as Table~\ref{tab:owt}. Duo has no public LM1B checkpoint and is not included.
}
\label{tab:lm1b}
\setlength{\tabcolsep}{4pt}\small
\begin{tabular}{l|ccc}
  \toprule
  Method & Gen.PPL($\downarrow$) & Entropy($\uparrow$) & MAUVE($\uparrow$) \\
  \midrule
  \midrule
  Real Text (LM1B) & 52.7$\pm$0.6 & 4.29$\pm$0.01 & 0.95$\pm$0.01 \\
  \midrule
  AR-B (re-trained) & \textbf{37.7$\pm$0.5} & 4.29$\pm$0.00 & 0.09$\pm$0.01 \\
  CoBit-S & 59.4$\pm$0.6 & 4.31$\pm$0.01 & --- \\
  LDLM$^{\dagger}$ & 63.0$\pm$0.5 & 4.37$\pm$0.00 & 0.91$\pm$0.02 \\
  T5Gemma-2 + ELF-B & 60.6$\pm$0.3 & 4.32$\pm$0.01 & 0.36$\pm$0.00 \\
  \textbf{Student (ours) + ELF-B} & \underline{50.5$\pm$0.5} & 4.31$\pm$0.00 & 0.36$\pm$0.04 \\
  \bottomrule
\end{tabular}
\end{table}

\subsection{Efficiency}

Scaling the embeddings is cheap for both training and sampling, as decomposed in Table~\ref{tab:cost}. For training, the embeddings are frozen and take up a small part of the FLOPs and latency, and can also be cached; our distilled student halves the encoding cost of T5Gemma-2. For sampling, they affect only the final decoding step by the LM head, which is fast and performed once per sample. This is why LM head parameters are less important and are put in a subscript in Table~\ref{tab:owt}. End-to-end, ELF-M produces a $1024$-token sequence in $1.5$\,s at NFE $64$ and $11.8$\,s at NFE $512$; GPT-2-M takes $10.7$\,s with $1024$ serial steps.

\begin{table}[!htb]\centering
\caption{\textbf{Cost of scaling the embeddings} per $1024$-token sequence (one H100, bf16, batch size $1$).}
\label{tab:cost}\setlength{\tabcolsep}{5pt}\small
\begin{tabular}{ll|ccc}
\toprule
 & & Params & FLOPs & Latency \\
\midrule\midrule
\multirow{5}{*}{\rotatebox{90}{\small Training}} & Encoding, T5-small & $19_{+16}$M & $0.04$\,T & $1.5$\,ms \\
 & Encoding, T5Gemma-2 & $100_{+168}$M & $0.21$\,T & $3.4$\,ms \\
 & Encoding, Student & $50_{+168}$M & $0.10$\,T & $2.1$\,ms \\
 & ELF-B denoiser step (fwd$+$bwd) & $89$M & $0.55$\,T & $11.2$\,ms \\
 & ELF-M denoiser step (fwd$+$bwd) & $327$M & $2.0$\,T & $30.6$\,ms \\
\midrule
\multirow{4}{*}{\rotatebox{90}{\small Sampling}} & ELF-B denoiser step & $89$M & $0.18$\,T & $11.8$\,ms \\
 & ELF-M denoiser step & $327$M & $0.67$\,T & $23.0$\,ms \\
 & Decoding, the denoiser pass & \multicolumn{3}{c}{one denoiser step as above} \\
 & Decoding, the $262$k LM head & $168$M & $0.34$\,T & $6.3$\,ms \\
\bottomrule
\end{tabular}
\end{table}

%% file: Sections/conclusion.tex
In this paper, we explore scaling the text embeddings that continuous DLMs operate on, and find that scaling improves diffusibility. We further find that scaling itself is not enough, as strong embeddings can be hard for continuous diffusion to generate perfectly. We view diffusibility as generating valid and decodable embeddings from the same DLM, and propose a distillation that further enhances the scaled embeddings. Perhaps counterintuitively, we find that such distillation trades discriminative power of the embeddings for generation, and an embedding space where similar embeddings are more connected is better for diffusion.

Our results serve as a starting point for what this paper leaves open: few/one-step generation (given ELF trained on raw T5Gemma-2 under-integrates at low NFE), practical-scale models for real-world tasks such as QA, and adapting existing AR models directly into embeddings for DLMs (T5Gemma-2 is adapted from Gemma-3~\citep{team2025gemma}, and one would want to skip that step). 
In all, our findings suggest that designing a suitable embedding space of DLMs is as important as designing the diffusion, and also provide a promising and reliable way to scale continuous DLMs from their embeddings.

%% file: Sections/appendix.tex
\section{Additional Experiments}

\subsection{Statistical results on the generated embeddings}\label{app:census}
We generate $64$ sequences of $1024$ tokens per model with ELF-B at NFE $32$ and keep every token position where the decoded top-1 is below $0.6$: $9064$ tokens for the teacher and $2664$ for the student. For each, we measure the distance from the generated embedding to the nearest candidate word. The teacher's embeddings land far (mean $1.39$\,nn, median $1.38$), the student's close (mean $0.82$, median $0.81$). Per sequence, the teacher has $142$ uncertain positions, of which $109$ are invalid; the student $42$ and $12$. We only inspect uncertain positions, and there may be more invalid embeddings in the full generated sequences.

\begin{figure}[h]\centering
\includegraphics[width=0.9\linewidth]{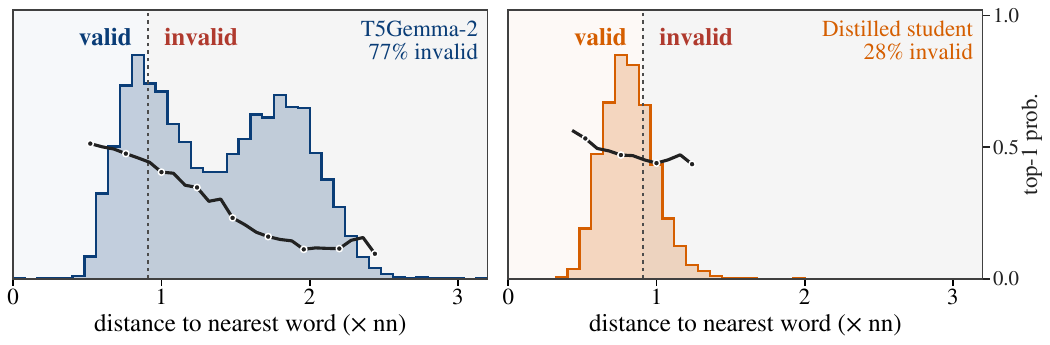}
\caption{\textbf{Student's and teacher's generated embeddings.} Positions with top-1 below $0.6$ from $64$ sequences per model (NFE $32$). Curve: the decoder's median top-1 per distance bin.
}
\label{fig:census}
\end{figure}

Fig.~\ref{fig:census} shows that a large embedding error means uncertain decoding: for the teacher, the median top-1 falls from $0.48$ within the $0.91$\,nn range to $0.23$ beyond it; for the student, it stays near $0.48$ and $0.45$. And the embedding error hurts the text quality: the $41$ (out of $64$) teacher sequences whose median distance is within $0.91$\,nn have Gen.\ PPL $18$--$106$, and the two beyond $1.8$\,nn read above $2000$; every student sequence has a median distance below $0.95$\,nn. 

\subsection{Candidates are robust to different decoders}\label{app:aragree}

We read the candidates of a token from the deployed decoding head, as it is the map from embeddings to words that the DLM uses, and the top-$k$ are the words it would decode to. To check that the decoder choice does not drive the analysis, we run the pretrained T5Gemma-2 decoder on the same sequences with teacher forcing. On real text, its top-$3$ shares at least one word with the head's at $99.3\%$ of $764$ positions ($1.5$ of $3$ on average). On generated text, the two decoders are unsure at the same positions: the entropy of their predicted word distributions correlates at $0.72$ over $3702$ positions. The pretrained decoder is less certain overall (median entropy $6.3$ nats), but it flags the same positions.

\subsection{Ablation on distillation}\label{app:ablation}

\begin{figure}[h]\centering

\includegraphics[width=.7\linewidth]{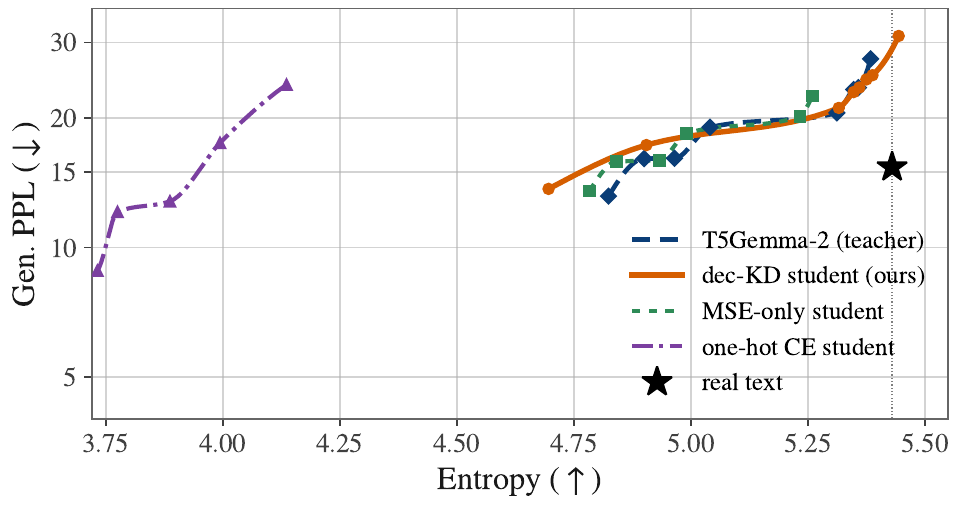}
\caption{
\textbf{Ablation on the distillation objective.} Gen.\ PPL--entropy curves on OWT-1024 with ELF-B. The teacher, the KL student and the MSE student are sampled under the same settings ($256$ samples, one seed), which vary the self-conditioning scale ($1$--$6$), the sampler's noise scale ($1.0$--$3.0$) and the NFE ($16$--$256$) but do not form a full grid; the CE student is sampled on the grid of Table~\ref{tab:owt}. Each curve is the best PPL at the same entropy. These settings do not contain the teacher's setting in Table~\ref{tab:owt} (sc$=1$, NFE$=64$), so in this plot the teacher stops at entropy $5.38$.
}
\label{fig:mse_front_n1024}
\end{figure}

\paragraph{Distillation objective.} 
We replace the distillation objective (KL on soft labels) with two other losses and train the same ELF-B, keeping everything else fixed. The first is an MSE between the student's and the teacher's embeddings, where no decoder is needed. It tests whether copying the teacher's embeddings is enough. The second is a cross-entropy on the input tokens (one-hot, hard labels) through the same frozen decoder. It tests whether the gain comes from training through the decoder, or from the soft labels. As shown in Fig.~\ref{fig:mse_front_n1024}, the MSE student reaches an entropy of only $5.26$, and the CE student stays below $4.18$. The raw teacher also stays below real-text entropy. In contrast, the student distilled with soft labels surpasses the teacher (similar to the recent on-policy distillation techniques~\citep{liang2026visual, wang2026look}) and achieves a Gen.\ PPL of $31.2$ at a real-text entropy of $5.44$. So it is the soft labels that make the embeddings diffusible.

\begin{figure}[h]\centering
\includegraphics[width=.7\linewidth]{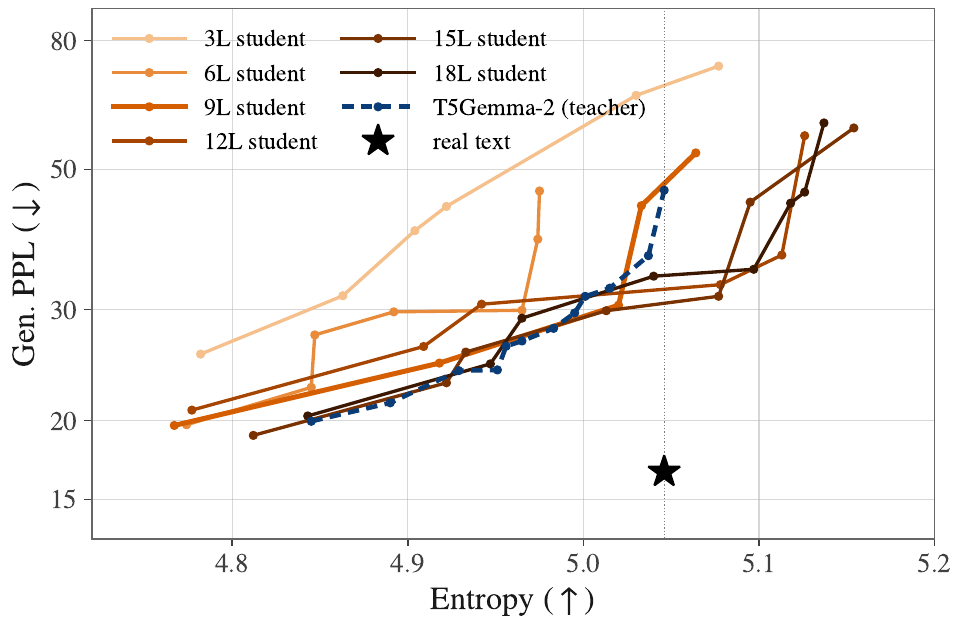}
\caption{\textbf{Effect of student depth.} Gen.\ PPL--entropy curves for students with $3$ to $18$ layers and the T5Gemma-2 teacher on the $512$-token OWT setting. All students use the same distillation and evaluation protocol. The vertical dotted line marks the entropy of real text, and the black star shows the real-text reference. Deeper students generally provide a better quality--diversity trade-off, although the improvement is not monotonic beyond $12$ layers.
}
\label{fig:ladder_fronts}
\end{figure}

\paragraph{Student depth effect.} We vary the student's depth from $3$ to $18$ layers to see whether the gain comes from the smaller capacity of the student. Everything else is fixed, and Fig.~\ref{fig:ladder_fronts} shows the Gen.\ PPL--entropy curves. Depth matters most near real-text entropy: at $5.05$, students with $12$--$18$ layers read $31$--$34$ Gen.\ PPL, the $9$-layer student $49$, and the teacher $46$; $3$ and $6$ layers fall behind. At lower entropy, the curves are within about $2$ PPL. The $18$-layer student is as deep as the teacher and still beats it, so the gain does not come from fewer layers. We use $9$ layers to halve the encoding cost. A randomly initialized $9$-layer student does not converge.

\subsection{How distillation changes the embeddings}\label{app:downstream}\label{app:realemb}

In Sec.~\ref{subsec:easier} we have shown that the embeddings generated on the student contain less error and decode back to tokens more accurately. Here we examine more directly how the distillation changes the embeddings, without the diffusion. We take $317$ positions from held-out OWT sequences. At each position we replace the word with its top-$2$ candidate, encode the sequence again, and measure the distance between the two embeddings at this position, in nn of each encoder.

\begin{figure}[!h]
\centering
\includegraphics[width=0.5\linewidth]{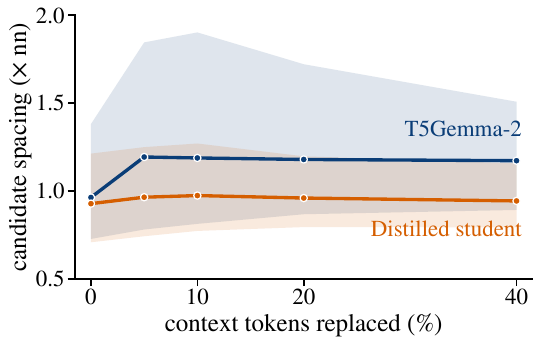}
\caption{\textbf{Distance between a word and its candidate on real text,} when a share of the context tokens is replaced. Line: median over $317$ positions; band: quartiles.}
\label{fig:realemb}
\end{figure}

At generation time, each token position sees a context that is still under-sampled and partly wrong. Here we imitate such a context on real text and without the diffusion: we replace a share of the context tokens with tokens from another document, and measure the distance between a word and its candidate (Fig.~\ref{fig:realemb}). Any difference then comes from the encoder alone. On clean text, the two encoders are close ($0.96$ vs.\ $0.93$\,nn). With $10\%$ of the context replaced, the candidates of T5Gemma-2 move apart to $1.19$\,nn, while the student's stay at $0.97$. The student keeps the candidates close when the context is imperfect. We leave a more fine-grained analysis of the geometry for future work.

\paragraph{Downstream performance of the distilled embeddings.}
Although the student is distilled only on OWT, it is still able to encode and decode other text. We test this on downstream tasks, similar to previous work on image diffusion~\citep{li2026evaluating}. We find top-$1$ token accuracy is $99.3$ for both teacher and student on WikiText-103~\citep{merity2016pointer}, and $99.3$ vs.\ $99.4$ on held-out arXiv papers~\citep{clement2019use} (Table~\ref{tab:tradeoff}). Discriminative power drops a little. With the encoder frozen, a linear probe on the mean-pooled tokens scores lower on SST-2, STS-B and MRPC (ridge regression for STS-B). 

\begin{table}[!h]
\centering
\caption{\textbf{Distillation trades discriminative power for diffusibility.} Reconstruction and probing results are reported on a $0$--$100$ scale. Gray values indicate generation results below the real-text entropy.}
\label{tab:tradeoff}
\setlength{\tabcolsep}{5pt}
\small
\begin{tabular}{l|c|cc|ccc}
\toprule
& Generative & \multicolumn{2}{c|}{OOD reconstruction} & \multicolumn{3}{c}{Discriminative} \\
Encoder ($+$ELF-M) & Gen.\ PPL@Entropy ($\downarrow$) & WikiText-103 & arXiv & SST-2 & STS-B & MRPC \\
\midrule
T5Gemma-2 & $19.3@5.44$ & $99.3$ & $99.3$ & $89.4$ & $71.7$ & $71.6$ \\
\textbf{Student (ours)} & $\mathbf{17.8}@5.45$ & $99.3$ & $99.4$ & $78.1$ & $65.3$ & $70.8$ \\
T5-small & \textcolor{gray}{$32.8@5.36$} & $100.0$ & $100.0$ & $84.4$ & $74.3$ & $74.8$ \\
\bottomrule
\end{tabular}
\end{table}

\section{Experimental details}\label{app:details}

We build on a public PyTorch implementation of ELF.\footnote{\url{https://github.com/Ugness/ELF-pytorch}} Unless otherwise specified, we follow the original training and sampling configuration as in ELF.

\subsection{Training Details}\label{app:train}

\paragraph{Main experiments.} We pack OWT~\citep{Gokaslan2019OpenWeb} into sequences of up to $1024$ tokens, each ending with an EOS, and obtain $9.0$M sequences ($9.2$B tokens) from the $7.8$M documents. Following exactly ELF's design, we train ELF-B and ELF-M with $x$-prediction flow matching, logit-normal time sampling, and learned prefix tokens that carry the self-conditioning scale, for $5$ epochs at batch size $512$, using Muon optimizer with learning rate $2\times10^{-3}$, $2$k warm-up steps, and EMA $0.9999$; we evaluate the EMA weights. On LM1B, ELF-B trains on $7.9$M $128$-token sequences for $5$ epochs with the same recipe.

\paragraph{Encoder comparison.} To compare different pretrained embedding models at lower computational cost, we train ELF-B on a $25\%$ subset of OWT using $512$-token sequences. This setting is used for the encoder comparison in Fig.~\ref{fig:panel}, including T5-small and T5-base~\citep{raffel2020exploring}, T5Gemma-1~\citep{zhang2025encoder}, T5Gemma-2~\citep{zhang2025t5gemma}, and ModernBERT~\citep{warner2025smarter}. 
We train on $2$M OWT documents, truncated to $512$ tokens (shorter ones padded), for $15$ epochs ($58.6$k steps) with the same recipe. The real-text reference at this length is $16.6@5.05$ ($15.4@5.43$ at $1024$ tokens), so we compare encoders only within this setting.

\subsection{Distillation details}\label{app:distill}

We initialize the $9$-layer student encoder using $9$ evenly spaced layers of the $18$-layer T5Gemma-2~\citep{zhang2025t5gemma} teacher. Only the student encoder is trained, while the others (teacher encoder and decoder, vocabulary embedding table) are frozen. For each training sequence, the teacher decoder is teacher-forced on the teacher encoder outputs to produce a token distribution at each position. We retain the teacher's top-$256$ logits and renormalize them to form the soft target distribution. The student representations are passed through the same frozen decoder, and the student encoder is trained to minimize the token-wise KL divergence to the teacher distribution, implemented as cross-entropy over the retained indices, and we mask the padding positions. 

We distill the student for $50$k steps on the same packed $1024$-token OWT sequences used to train the diffusion models. We use AdamW with a learning rate of $3\times10^{-4}$, $1$k warm-up steps, and a batch size of $32$ per GPU across $8$ GPUs. We select the checkpoint with the lowest held-out distillation loss. Training takes $38.6$ hours on $8$ H100 GPUs. After distillation, we train a separate ELF model on the student embedding space using the identical architecture and training configuration as the one trained on the original T5Gemma-2 embeddings.

\subsection{Evaluation Details}\label{app:eval}

\paragraph{Metrics.} We re-tokenize every generated and real-text sample using the GPT-2-Large tokenizer~\citep{radford2019language}. We compute Gen.\ PPL with GPT-2-Large and unigram entropy under the same tokenization. The held-out OpenWebText (OWT) reference~\citep{Gokaslan2019OpenWeb} has a Gen.\ PPL of $15.4$ and an entropy of $5.43$. For MAUVE~\citep{pillutla2021mauve}, we use the GPT-2-Large featurizer with $256$-token features, $1024$ generated samples, and a fixed held-out reference set of the same size. The real-text MAUVE value is computed between disjoint held-out splits and is below $1$ because the metric is estimated from finite samples.

\paragraph{Sampling and operating-point selection.} 
We use the SDE sampler of ELF~\citep{hu2026elf} with churn $\gamma=1.5$ and noise scale $2.0$. NFE means the number of denoiser evaluations. For each embedding (T5Gemma-2 or the student) and each model size (ELF-B or ELF-M), we sample on the same grid: self-conditioning scale (sc) in $\{1,2,3\}$ $\times$ NFE in $\{64,128,256,512\}$, with $256$ samples and seed $0$ (Fig.~\ref{fig:sampling_abl}). We pick the point with the lowest Gen.\ PPL among those that reach the real-text entropy $5.43$. If none reaches it, we report the point with the highest entropy and mark the row in gray. The chosen point is then evaluated on three seeds with $1024$ samples each, and we report the mean and standard deviation. 
The chosen points in Table~\ref{tab:owt} are:
\begin{center}\small\setlength{\tabcolsep}{8pt}
\begin{tabular}{l|cc}
\toprule
Model & sc & NFE \\
\midrule
T5Gemma-2 $+$ ELF-B & $1$ & $64$ \\
Student $+$ ELF-B & $1$ & $256$ \\
T5Gemma-2 $+$ ELF-M & $2$ & $256$ \\
Student $+$ ELF-M & $3$ & $512$ \\
\bottomrule
\end{tabular}
\end{center}

sc acts as a guidance scale on the self-conditioning input, and a larger sc promotes iterative refinements during sampling, promoting consistency and fluency but also sacrificing diversity. So we see that in Fig.~\ref{fig:sampling_abl}, with larger sc, the entropy and Gen.\ PPL are also smaller. This is similar to the classifier-free guidance~\citep{ho2022classifier} in image diffusion models, where it improves quality but degrades diversity~\citep{li2026mclr}.

We do not use NFE below $64$ in the main evaluation, since the samples collapse there: T5Gemma-2$+$ELF-B reads Gen.\ PPL $77$--$335$ at NFE $16$--$32$, and T5-small$+$ELF-M $42$--$84$. In the teaser (Fig.~\ref{fig:opening}), we still include these points to give the full PPL--entropy curve. This is why T5-small$+$ELF-M passes real-text entropy in the teaser but is gray in Table~\ref{tab:owt}.

\paragraph{Baseline evaluation protocols.} 

Rows with $^{\dagger}$ are copied from the papers, under their own protocols, so they are not directly comparable. All other rows are our models or public checkpoints, and we sample with the authors' code and score with $3\times1024$ samples. Duo~\citep{sahoo2025diffusion} uses $T=512$ and nucleus sampling at $p=0.94$; CoBit~\citep{batzolis2026cobit} uses its released sampler, with NFE $256$ for CoBit-S and $160$ for CoBit-M. GPT-2-S and GPT-2-M are sampled at temperatures $0.925$ and $0.885$, where they reach real-text entropy. Parameters are counted at generation time as trunk$_{+\mathrm{head}}$; the head is the token embedding table and the output projection (for CoBit, its bit projections). For AURORA-LM, we count the denoiser alone, since its autoencoder is not public. CoBit-S does not reach real-text entropy on OWT, so we report its best Gen.\ PPL within $0.1$ of it ($37.3\pm0.5$ at $5.33$). 

On LM1B, every point of the OWT grid gives samples above real-text entropy ($4.29$). So we extend the grid to sc $\{4,5,6,8\}$ at NFE $\{128,256\}$ for both encoders and apply the same rule, and the best sampling configuration for both is sc$=6$/NFE$=256$. There is no common AR baseline on LM1B, so we train AR-B, a $12$-layer GPT-2-style model, from scratch on the same data (T5Gemma-2 tokens, $5$ epochs) and sample it at temperature $0.8$. CoBit-S is scored under its authors' protocol (uncased WordPiece text, entropy over BERT ids), which reproduces their number ($59.4\pm0.6$; reported $59.8\pm0.6$); we leave its MAUVE blank, since uncased text cannot be compared with our cased references. LDLM rows are the authors' numbers (OWT at NFE $256$, LM1B at NFE $512$).

\paragraph{Tokenization.}\label{app:tokeq}
One problem with the evaluation is that the encoders use different tokenizers. For instance, T5Gemma-2 uses the $262$k multilingual Gemma vocabulary~\citep{team2025gemma}, T5-small uses a $32$k SentencePiece vocabulary~\citep{raffel2020exploring}, GPT-2 uses a $50$k BPE~\citep{radford2019language}, so a $1024$-token sequence covers a different amount of text in each. Here, we verify that the improvements are not hacked by using different tokenizers or generating texts of longer effective lengths.

To make the methods comparable, we re-tokenize both generated and real text with the GPT-2-Large tokenizer before computing Gen.\ PPL and unigram entropy. At the sampling configurations used in the main evaluation (Table~\ref{tab:owt} and Fig.~\ref{fig:sampling_abl}), the teacher and student samples contain median lengths of $1021$ and $1027$ GPT-2-Large tokens per window, respectively, compared with $1008$ for real text. Their character counts are also similar ($4.7$k and $4.8$k versus real $4.5$k), and non-ASCII characters account for $0.3\%$ in all three. We also find that truncating real windows by $5\%$ changes unigram entropy by only $0.031$, suggesting that the observed $1$--$2\%$ difference in token length contributes $<0.01$ to the reported entropy.

\begin{table}[h]
\centering
\caption{\textbf{Sensitivity of MAUVE to feature length.} We vary only the maximum feature length while keeping the generated samples, reference samples, and GPT-2-Large featurizer unchanged. Increasing the feature length substantially lowers the scores of all models. We report three decimal places to distinguish the small scores and standard deviations in this analysis.}
\label{tab:mauve_sensitivity}
\setlength{\tabcolsep}{8pt}
\small
\begin{tabular}{lcc}
\toprule
& \multicolumn{2}{c}{MAUVE ($\uparrow$)} \\
& $256$-token features & $1024$-token features \\
\midrule
Real vs.\ real & $.944$ & $.924$ \\
\midrule
GPT-2-S & $.878\pm.020$ & $.183\pm.026$ \\
T5-small $+$ ELF-B & $.415\pm.037$ & $.012\pm.004$ \\
T5-small $+$ ELF-M & $.858\pm.017$ & $.016\pm.001$ \\
T5Gemma-2 $+$ ELF-B & $.758\pm.052$ & $.103\pm.006$ \\
Student $+$ ELF-B & $.856\pm.014$ & $.111\pm.009$ \\
\midrule
GPT-2-M & $.911\pm.010$ & $.169\pm.015$ \\
T5Gemma-2 $+$ ELF-M & $.861\pm.023$ & $.098\pm.006$ \\
Student $+$ ELF-M & $.889\pm.013$ & $.109\pm.008$ \\
\bottomrule
\end{tabular}
\end{table}

\input{Sections/appendix_defects}
\subsection{Embedding-space Visualization Details}\label{app:maps}

Each embedding panel in Fig.~\ref{fig:opening}, Fig.~\ref{fig:crystal} and Fig.~\ref{fig:student_mitigation} visualizes a single token position. We show the generated embedding produced by the diffusion model together with the candidate embeddings associated with the decoder's top-$3$ predicted tokens at that position. All distances are first computed in the full $640$-dimensional embedding space and then normalized by the median nearest-neighbor distance between real embeddings in that space, denoted by $\mathrm{nn}$. This reference distance is $13.5$ for the teacher and $12.4$ for the student. A distance of $1\,\mathrm{nn}$ therefore represents the typical spacing between neighboring real embeddings, allowing teacher and student panels to be compared relative to the local scale of each space. The bars show the decoder's probabilities. Decoding is nonlinear, so the most probable word is not always the geometrically closest candidate.

We obtain the plotted coordinates using weighted metric multidimensional scaling with SMACOF \citep{de2009multidimensional}. In the dimension reduction, the distances between the generated embedding and each candidate get weight $2.5$, since these are the distances we want preserved. Across the displayed panels, the mapped generated-to-candidate distances differ from their full-dimensional values by at most $13\%$. We use a distance-preserving map rather than a linear projection such as PCA because the embedding space is still highly nonlinear, and the generated embeddings can lie far outside the subspace spanned by the candidate embeddings; a projection is then not meaningful.

\section{Generated Examples}\label{app:examples}

We provide qualitative examples from OWT-1024 and LM1B under different sampling configurations. They illustrate the generation behavior of ELF~\citep{hu2026elf} models trained on the raw T5Gemma-2~\citep{zhang2025t5gemma} and our distilled student embeddings. For each sample, seq.\ PPL is its GPT-2-Large perplexity and $H$ its (unigram) entropy as we report throughout the paper, both over the sample's first $1024$ GPT-2 tokens. At low NFE (e.g., $64$ here), some T5Gemma-2 samples collapse~\citep{shi2026closer} into a funny language mix since the Gemma vocabulary is multilingual,\footnote{The GPT-2-Large judge also scores non-English text, so a mixed-language sample can still lead to a low PPL. The boxes show excerpts, and the rest of these two samples is mostly English.} while the distilled student samples stay fluent.

\subsection{OWT-1024}

\begin{teacherbox}{T5Gemma-2 $+$ ELF-M (sc$\,2$/NFE$\,64$) \hfill seq.\ PPL $19.4$, $H$ $5.13$}
\includegraphics[width=\linewidth]{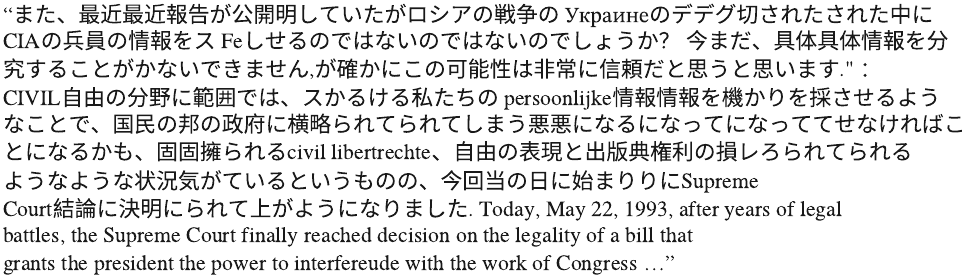}
\end{teacherbox}

\begin{teacherbox}{T5Gemma-2 $+$ ELF-M (sc$\,2$/NFE$\,64$) \hfill seq.\ PPL $24.2$, $H$ $5.48$}
\includegraphics[width=\linewidth]{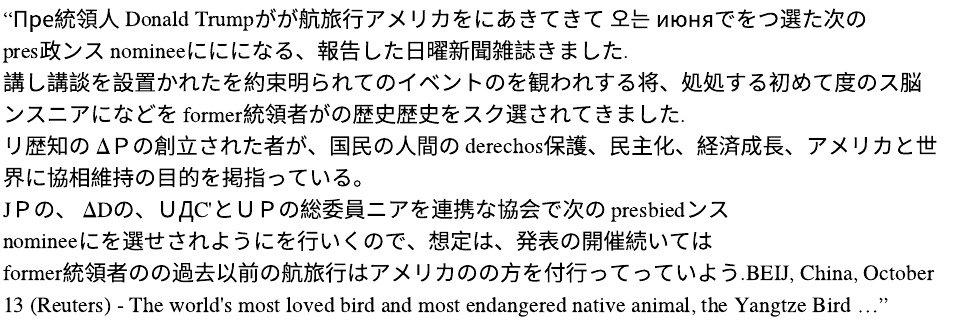}
\end{teacherbox}

\begin{teacherbox}{T5Gemma-2 $+$ ELF-M (sc$\,2$/NFE$\,256$) \hfill seq.\ PPL $23.0$, $H$ $5.46$}
``A series of advanced radio telescopes have been studying the cold atmospheric
surface of a star called Cepulus, which lies 1.4 billion light years from our
sun. A traditional astronomical theory is that the star's cold atmosphere is a
result of gravitational activity and radiation.

But two University of Michigan astronomers, both co-authors of a new paper
published online this week in the Proceedings of the National Acad Society,
suggest that the star is made up of two different types of clouds. One of
these clouds contains tiny membranes similar to those that attach to our own
ears, a portion of our vision system. The other cloud contains a small number
of smaller splles smaller than a baseball ball at the end of a baseball
field.

The findings published in the June issue of Astronomomy suggest the clouds
may have been formed in response to radiation, but the star is very close to
its parent star, which makes it even further away from Earth\ldots''
\end{teacherbox}

\begin{teacherbox}{T5Gemma-2 $+$ ELF-M (sc$\,2$/NFE$\,256$) \hfill seq.\ PPL $18.2$, $H$ $5.46$}
``Decades of research have shown that deep-fracking has led to the release of
toxic methane and carbon dioxide into the water system during the drilling
processes used to explore a well's depth. The administration's proposed
measure would ban methane, carbon dioxide and other components of oil and gas
from appearing in drinking water for at least five years. It would also
require state regulators to regulate toxic materials in the use of drinking
water.

The nation's oil and gas industry generates about \$1 trillion in revenue and
costs about \$400 billion a year. Deep-fracking releases more than 11,000
cancer-threataging chemicals and about 10,000 tons of carbon dioxide.
Toxicive chemicals are linked to health risks, including major health
problems such as heart disease, immune immune disease, diabetes, diabetes,
leukemia and cancer\ldots''
\end{teacherbox}

\begin{studentbox}{Student $+$ ELF-M (sc$\,3$/NFE$\,64$) \hfill seq.\ PPL $14.2$, $H$ $5.51$}
``The crash took place at an intersection near East 8th Street and Virginia
Avenue. Police and law enforcement officers with the Chicago CHP Unit located
the driver driving the vehicle.

The suspect was identified as 27-year-old Michael Pattiani. He is charged
with misdemeanor use of carrying a firearm, drug trafficking, illegal
possession of controlled drugs, marijuana trafficking, and possession of one
or more amounts of marijuana, manufacturing and distribution of commercial
controlled drugs. The charges range from at least one month to life in
prison and are subject to a maximum fine of \$100,000\ldots''
\end{studentbox}

\begin{studentbox}{Student $+$ ELF-M (sc$\,3$/NFE$\,256$) \hfill seq.\ PPL $11.8$, $H$ $5.47$}
``It's also interesting to note that Sanders is probably not the only guy
around when it comes to fighting global warming. The Clinton campaign hasn't
dropped a single word on the issue of climate change to date, and the
popularity of the Sanders campaign's ``skip the wall'' rhetoric makes it
clear how little the campaign has actually done on the issue. In the wake of
this election, the results of recent exit polls have shown that both sides of
the political spectrum have realized the need to call out each other with
bogus statements before they can even get to the polls.

If you're ready to read more from the unbossed and unbought Politicus team,
sign up for our newsletter here! Email address: Leave this field empty if
you're human:At least four people were killed in an airstrike in eastern
Syria on Sunday, coming just days after the U.N. Security Council passed a
resolution in support of ending the three-year civil war\ldots''
\end{studentbox}

\begin{studentbox}{Student $+$ ELF-M (sc$\,3$/NFE$\,256$) \hfill seq.\ PPL $11.6$, $H$ $5.42$}
``In June, Sunpay drew fire when it announced a new technology, ShadowPass, which offered the ability to encrypt phone calls and text messages without requiring a password or consent. The revelations of ShadowPass's encryption techniques put several ISPs, such as AT\&T Broadband Co, Verizon Wireless Co, Time Warner Co, T-Mobile Co and Sunpay, under the microscope for alleged fraud, identity theft and data leaking.

In October, Sunpay reached a settlement with the U.S. Justice Department in New York for violating secret orders from the FBI. The company argued that even though it had agreed to comply with enforcement and information-sharing regulations, the Justice Department went too far and ordered the company to pay settlement costs and damages. The FBI later alleged that Sunpay used similar encryption techniques to ShadowPass in order to share customer data with foreign intelligence agencies.''
\end{studentbox}

\subsection{LM1B}

\begin{teacherbox}{T5Gemma-2 $+$ ELF-B (sc$\,6$/NFE$\,256$) \hfill PPL $32.5$, $H$ $4.29$}
``The well-known Illinois senator spent more than 12 years of working with the White
House under President George W. Bush from 1991-98 , was deputy economic adviser for
President Bush and had a key role in the beginning of President Bush 's administration
.A US military official says the Taliban killed four British soldiers in a car attack
at a military base in southern Afghanistan .Sales for the year rose by 24.4 \% 2009
after a 5.8 \% rise in 2008 , but August sales figures were far lower than many
economists 's expectations\ldots''
\end{teacherbox}

\begin{teacherbox}{T5Gemma-2 $+$ ELF-B (sc$\,6$/NFE$\,256$)\hfill PPL $60.3$, $H$ $4.31$}
``Other improvements at TG include the Interactive Television and Video Matching
System ( IMC ) , the Interactive Abreel Recall Retrieval System ( IAC ) for file
matching technology , and the Integrated Video in Data Modification ( IMP ) system
that allows the television staff to watch television shows two to three hours .State
police spokesman Ron Parner said the police were still investigating Wright 's case
and that the agency had no evidence of any wrongdoing .In London , the British Bank
of England ( BCC ) said on Tuesday that the economy could lose 80 per cent of the
global economy in the second quarter and that unemployment would continue to rise .''
\end{teacherbox}

\begin{studentbox}{Student $+$ ELF-B (sc$\,6$/NFE$\,256$) \hfill PPL $30.1$, $H$ $4.32$}
``The cost-reduction plan will see not only reduce operating costs ; it is focused on
increasing costs and increasing staff loyalty to reduce the overall costs .In a
statement released Wednesday , the company said it would make a net loss of more than
\$ 100 billion under its bailout plan , but could not immediately say how much income
would be paid from the sale .A third man has died in the crash of a transport truck in
Bournemouth , a spokeswoman said .Mr Wood said that the deal would not be made by the
time of this year 's end , but that it was not clear whether he would sell the
shares\ldots''
\end{studentbox}

\begin{studentbox}{Student $+$ ELF-B (sc$\,6$/NFE$\,256$) \hfill PPL $50.4$, $H$ $4.36$}
``The cost of the digital screens rose more than half last year , rising from
\pounds 8m to \pounds 16m in 2007 .Arini Szymeva won a world-record 62 at women 's
Beijing Olympics last year , and in 2008 she was awarded \$ 69m in wins at the China
International Tennis Tournament .There are a variety of occasions in which you feel
like you 're getting love .LONDON , England ( AP ) - A man accused of killing a
British military officer in Afghanistan has been sentenced to five years and six
months , British police said Thursday .''
\end{studentbox}